\documentclass[11pt]{article}

\usepackage[final]{acl}

\usepackage{times}
\usepackage{latexsym}
\usepackage{needspace}
\usepackage[T1]{fontenc}

\usepackage[utf8]{inputenc}

\usepackage{microtype}

\usepackage{inconsolata}
\usepackage{tabularx}

\usepackage{graphicx}
\usepackage{amsmath}
\usepackage{amssymb}
\usepackage{booktabs}
\usepackage{enumitem}
\usepackage{algorithm}
\usepackage{algpseudocode}
\usepackage{multirow}
\usepackage{float}
\usepackage{stfloats}
\usepackage{xspace}
\usepackage{placeins}
\usepackage{tcolorbox}
\usepackage{listings}
\tcbuselibrary{skins,listings}
\tcbset{
  basebox/.style={
    fonttitle=\footnotesize\bfseries\sffamily,
    enhanced, boxrule=0.6pt, arc=2pt,
    left=5pt, right=5pt, top=3pt, bottom=3pt,
    toptitle=2pt, bottomtitle=2pt
  }
}
\newtcolorbox{specbox}{basebox,
  title=Degraded Spec,
  colback=gray!8, colframe=gray!55, colbacktitle=gray!22,
  coltitle=black
}
\newtcolorbox{askbox}[1]{basebox,
  title=#1,
  colback=orange!7, colframe=orange!55, colbacktitle=orange!22,
  coltitle=black
}
\newtcolorbox{simbox}{basebox,
  title=Simulator,
  colback=blue!7, colframe=blue!40, colbacktitle=blue!18,
  coltitle=black
}
\newtcolorbox{answerbox}[1]{basebox,
  title=#1,
  colback=green!6, colframe=green!45, colbacktitle=green!20,
  coltitle=black
}
\newtcolorbox{promptbox}[1]{basebox,
  title=#1,
  colback=gray!8, colframe=gray!55, colbacktitle=gray!22,
  coltitle=black
}

\newcommand{\passat}{\mbox{pass@1}\xspace}
\newcommand{\avgq}{\bar{q}}
\newcommand{\dbudget}{d_1}

\title{
  Pareto-Dominant Clarification:
  Post-Training Coding LLMs via PPO-Lagrangian Budget Constraints
  \textsuperscript{\(\dagger\)}
}

\author{
  Abhinav Rajput  \textsuperscript{*} \\ 
  Center for Data Science \\
  New York University \\
  New York, NY, USA \\
  \texttt{abhinav.rajput@nyu.edu}
  \And
  Acey Vogelstein \textsuperscript{*} \\
  Center for Data Science \\
  New York University \\
  New York, NY, USA \\
  \texttt{av3848@nyu.edu}
}

\begin{document}

\maketitle

\begingroup
\renewcommand{\thefootnote}{}

\footnotetext{
  \textsuperscript{\(\dagger\)} Accepted for the 2026 LMP Workshop,
  held in conjunction with the Conference on \textit{Empirical Methods in
  Natural Language Processing} (EMNLP).
}

\footnotetext{
  \textsuperscript{*} Equal contribution.
}

\endgroup

\begin{abstract}
Coding agents operating under ambiguous instructions or user prompts must decide whether to ask clarifying questions or attempt a solution directly. While clarification from the user may improve the correctness of the agent’s solution, each back-and-forth interaction incurs user and system costs, forming an explicit accuracy vs. efficiency tradeoff. Existing works study clarification behavior but do not train policies under enforceable clarification budgets; penalty-based approaches typically require separate coefficient tuning swept across all clarification budget levels. We formulate clarification as a Constrained Markov Decision Process (CMDP) and post-train Qwen2.5-Coder-7B-Instruct with PPO-Lagrangian to optimize coding accuracy, subject to an expected question-budget constraint. Evaluated on HumanEvalComm with a GPT-4o-mini oracle simulator, the resulting policies reveal that untuned clarification behavior is Pareto-inefficient: budget-constrained policies can simultaneously achieve higher accuracy and lower clarification rates than the baseline model. Across budget levels, we observe a log-shaped Pareto frontier with diminishing returns to additional clarification. Gains arise not from simply asking more questions overall, but from improved question targeting and better code generation under ambiguity. Without explicit supervision, trained policies learn to allocate clarification budget non-uniformly, asking more frequently on tougher (multi-degradation) tasks. These results suggest that unconstrained interactive LLM systems may systematically use clarification inefficiently.
\end{abstract}

\section{Introduction}
\label{sec:intro}

Coding agents given ambiguous, incomplete, or inconsistent task specifications must decide at each turn: ask a targeted clarifying question to recover missing information, or submit code directly and risk failure.
Asking can be decisive on genuinely underspecified tasks, but each clarification turn costs user attention and system resources.
Learning \emph{when} to ask (and, jointly, \emph{what} to ask) under an explicit, enforceable budget on expected clarification cost is the problem we address. Because a single policy generates both the routing decision and the clarifying question, our method optimizes the two together rather than treating question content as fixed.

Prior work on clarifying questions~\citep{rao-daume-iii-2018-learning, DBLP:journals/corr/abs-1907-06554, Wu_2025, curuksu2024policy} and constrained RL and LLM alignment~\citep{Altman1999ConstrainedMD, pmlr-v119-stooke20a, ouyang2022traininglanguagemodelsfollow} does not train the ask-vs-answer routing policy (the decision at each turn whether to ask a clarifying question or submit code) end-to-end under an explicit per-episode clarification-cost budget.
The CMDP framing naturally accommodates arbitrary budgets within a single training run, whereas penalty-based approaches require per-budget coefficient tuning.

We formalize the ask-vs-answer decision as a \textbf{Constrained Markov Decision Process} (CMDP;~\citealt{Altman1999ConstrainedMD}), in which the agent maximizes \textbf{\passat} (defined as the average fraction of unit test cases passed per problem, a continuous score in $[0,1]$) subject to a budget $\dbudget$ on the expected per-episode question count.
We post-train \textbf{Qwen2.5-Coder-7B-Instruct}~\citep{hui2024qwen25codertechnicalreport} with LoRA~\citep{hu2021loralowrankadaptationlarge} via \textbf{PPO-Lagrangian}~\citep{schulman2017proximalpolicyoptimizationalgorithms, Altman1999ConstrainedMD, pmlr-v119-stooke20a}, whose dual-ascent Lagrange multiplier enforces the budget without per-budget penalty tuning.
We evaluate policies in $(\avgq, \passat)$ space; a policy \textbf{Pareto-dominates} another if it achieves strictly higher \passat \emph{and} strictly lower $\avgq$ simultaneously.
Our contribution is not a new RL algorithm but the end-to-end, budget-constrained post-training of the ask-vs-answer routing policy: the CMDP formulation (Eq.~\eqref{eq:objective}) is what defines the problem, and other constrained-RL optimizers could target the same formulation.

\paragraph{Contributions.}
\begin{itemize}[noitemsep,topsep=2pt,leftmargin=*]
  \item We show that untuned LLMs may have Pareto-inefficient clarification policies: PPO-Lagrangian training simultaneously achieves higher accuracy \emph{and} lower question asking rates, correcting the baseline's operating point within a single training paradigm.
  \item Empirical Pareto analysis across seven budget levels reveals a log-shaped frontier with an efficient operating point: beyond a budget threshold, the friction cost of additional clarifications outweighs their marginal accuracy benefit.
\item Two structural findings help explain the gains: policies implicitly
learn (1) \emph{when to ask}, allocating their question budget non-uniformly
without explicit supervision and asking more on harder multi-degradation
problems, and (2) \emph{what to ask}, improving question quality and code
generation under ambiguity.
\end{itemize}
The complete implementation and trained checkpoints are available at
\href{https://github.com/Abhinav0710rajput/coding_llms_cmdp}{{our GitHub repository}}.

\section{Related Work}
\label{sec:related}

\paragraph{Clarifying questions in NLP.}
\citet{rao-daume-iii-2018-learning} rank clarification candidates via neural expected value of perfect information on StackExchange data.
\citet{DBLP:journals/corr/abs-1907-06554} study clarifying questions in open-domain information-seeking dialogue.
\citet{Wu_2025} benchmark LLM asking behaviour on degraded coding specifications without training a budget-constrained policy.
\citet{curuksu2024policy} trains a clarification fallback policy via PPO in multi-turn orchestration environments, but without an explicit per-episode budget constraint; their reward is defined over tool/intent matching rather than task correctness.
Recent work studying uncertainty-aware and learned clarification: \citet{zhang2023clarifynecessaryresolvingambiguity} and \citet{suri2026structureduncertaintyguidedclarification} estimate when clarification is useful, while the latter also uses RL, but neither trains under explicit clarification-cost constraints. \citet{sun2025trainingproactivepersonalizedllm} uses multi-objective RL with interaction costs incorporated through reward shaping rather than constrained optimization.
In coding, \citet{darji2025curiositydesignllmbasedcoding} trains clarification components, while \citet{vijayvargiya2026ambigsweinteractiveagentsovercome} evaluates interactive agents under underspecified tasks; neither trains a budget-constrained clarification policy.

\paragraph{Constrained RL and alignment.}
Constrained MDPs~\citep{Altman1999ConstrainedMD} formalise budget-constrained optimisation; PID-Lagrangian dual updates~\citep{pmlr-v119-stooke20a} and reward-constrained policy optimisation~\citep{tessler2018rewardconstrainedpolicyoptimization} address stability in constrained RL.
Budgeted and cost-aware dialogue policies have likewise been framed as CMDPs~\citep{zhang2019budgeted, waki2025learning}; our contribution is not the CMDP framing itself but its application to end-to-end, budget-constrained post-training of the ask-vs-answer routing policy for multi-turn coding agents.
Reinforcement Learning from Human Feedback (RLHF;~\citealt{ouyang2022traininglanguagemodelsfollow}) aligns LLMs via human feedback without per-episode cost constraints; Safe RLHF~\citep{dai2023saferlhfsafereinforcement} extends this via a CMDP formulation with PPO-Lagrangian to trade off helpfulness against a harmlessness cost.


\section{Method}
\label{sec:method}

\subsection{CMDP Formulation}
\label{sec:cmdp}

We model the multi-turn clarification task as a CMDP~\citep{Altman1999ConstrainedMD}
$\mathcal{M} = (\mathcal{S}, \mathcal{A}, P, R, C_q, C_t, \dbudget, d_2)$.
Full formal definitions (state, transition function, reward) are in Appendix~\ref{sec:appendix}.

\paragraph{Actions.}
\begin{equation}
  a_t \;\in\; \bigl\{\,\texttt{[ASK]}\;q_t,\quad \texttt{[ANSWER]}\;c_t\,\bigr\},
  \label{eq:actions}
\end{equation}
where $q_t$ is a clarifying question and $c_t$ is Python code; routing is via prefix
log-probability scoring (Section~\ref{sec:interaction}).

\paragraph{Costs and budget.}
The primary cost $C_q(a_t)$ counts the number of distinct atomic questions in an \texttt{[ASK]} turn, as determined by the user simulator (details in Appendix~\ref{sec:appendix}); $C_q = 0$ for \texttt{[ANSWER]}.
Counting individual questions rather than turns prevents the agent from exploiting the budget by bundling multiple requests into a single \texttt{[ASK]}.
The secondary cost $C_t(a_t) = 1$ accrues per turn.
Budget $\dbudget$ constrains the question count per episode on average across the training distribution; $d_2 = 4$ provides a soft turn-count constraint below the hard cap $T_{\max} = 6$.

\begin{equation}
\pi^* = \arg\max_\pi \mathbb{E}_\pi[R] \ \text{s.t.}\ \mathbb{E}_\pi[C_i] \leq d_i,\ i \in \{q, t\}.
\label{eq:objective}
\end{equation}
\subsection{Interaction Protocol}
\label{sec:interaction}

At each turn the agent outputs \texttt{[ASK]} (passed to a GPT-4o-mini simulator,
which returns a factual answer) or \texttt{[ANSWER]} (code executed against hidden
tests; episode terminates). Termination conditions and routing details are in Appendix~\ref{sec:appendix}.

\subsection{PPO-Lagrangian Optimization}
\label{sec:ppolag}

\paragraph{Policy and value heads.}
The policy is Qwen2.5-Coder-7B-Instruct~\citep{hui2024qwen25codertechnicalreport} with LoRA~\citep{hu2021loralowrankadaptationlarge} ($r{=}16$, $\alpha{=}32$); architecture, value head design, and training objective details are in Appendix~\ref{sec:appendix}.
We solve Eq.~\eqref{eq:objective} via Lagrangian relaxation \citep{Altman1999ConstrainedMD, pmlr-v119-stooke20a}, approximated via a simultaneous primal-dual method over policy parameters and non-negative Lagrange multipliers $\lambda_1$ (question budget) and $\lambda_2$ (turn budget), with $\lambda_1, \lambda_2 \geq 0$.

\paragraph{Generalized Advantage Estimation~\citep{schulman2016highdimensionalcontinuous}.}
Three GAE streams for $X \in \{R, C_q, C_t\}$ (recurrence in Appendix~\ref{sec:appendix}) are whitened to zero mean and unit variance before combination:
\begin{equation}
  A^{\mathrm{lag}}_t = \tilde{A}^R_t - \lambda_1\,\tilde{A}^{C_q}_t - \lambda_2\,\tilde{A}^{C_t}_t.
  \label{eq:lag_adv}
\end{equation}
The policy is updated via the clipped PPO objective~\citep{schulman2017proximalpolicyoptimizationalgorithms} ($\varepsilon{=}0.2$) over continuation tokens only, with prefix tokens excluded from likelihood ratios (equation in Appendix~\ref{sec:appendix}).
\paragraph{Dual update.}
After each rollout batch, and \emph{before} the PPO gradient step:
\begin{align}
  \lambda_1 &\leftarrow \operatorname{clip}\!\bigl(\lambda_1 + \eta_{\lambda_1}(\avgq - \dbudget),\;
    0,\; \lambda_{1,\max}\bigr), \label{eq:dual1} \\
  \lambda_2 &\leftarrow \operatorname{clip}\!\bigl(\lambda_2 + \eta_{\lambda_2}(\bar{C}_t - d_2),\;
    0,\; \lambda_{2,\max}\bigr), \label{eq:dual2}
\end{align}
where $\bar{C}_t$ is the mean turn count per rollout batch; $\eta_{\lambda_1}{=}0.1$, $\eta_{\lambda_2}{=}0.005$; $\lambda$ bounds in Table~\ref{tab:hyperparams}.
Training setup, checkpoint selection criteria, and Pareto frontier construction are detailed in Appendix~\ref{sec:appendix}.

\section{Experimental Setup}
\label{sec:setup}

\paragraph{Benchmark and evaluation metric.}
We train and evaluate on HumanEvalComm~\citep{Wu_2025}, a multi-turn
extension of HumanEval~\citep{chen2021evaluatinglargelanguagemodels} in which each problem
specification has been deliberately degraded before being presented to the model.\footnote{Seven degradation types: \texttt{1a}, \texttt{1c}, \texttt{1p}, \texttt{2ac}, \texttt{2ap}, \texttt{2cp}, \texttt{3acp}; numeric prefix = number of simultaneous degradations; letter suffix = type (\textbf{a}~=~ambiguity, \textbf{c}~=~contradiction/inconsistency, \textbf{p}~=~partial truncation).}
Each episode presents the model with a degraded specification; the model may ask clarifying questions (answered by a GPT-4o-mini simulator from the ground-truth oracle) before submitting a Python solution. Every instance presented to the model is degraded; HumanEvalComm contains no unmodified specifications.
Our primary evaluation metric is \passat\ (defined in Section~\ref{sec:intro}).
We split 772 degraded instances into 303 for training, 52 for validation, and 417 for held-out testing (split details in Appendix~\ref{sec:appendix}).

\textbf{Policy and simulator.} We fine-tune Qwen2.5-Coder-7B-Instruct~\citep{hui2024qwen25codertechnicalreport} using LoRA~\citep{hu2021loralowrankadaptationlarge}\footnote{Artifact licenses: HumanEval is released under the MIT License; Qwen2.5-Coder-7B-Instruct and HumanEvalComm are released under Apache-2.0. The GPT-4o-mini simulator is accessed via the OpenAI API under its terms of use. All artifacts are used in a manner consistent with their intended research use.}; both the policy and simulator receive a system prompt (full text and rationale in Appendix~\ref{sec:appendix_prompts}; representative episodes in Appendix~\ref{sec:appendix_episodes}).
We train 10 policies at seven budget levels, with two independent runs (different seeds, otherwise identical) at $\dbudget \in \{0.5, 0.75, 1.0\}$ to measure run-to-run variance.
The baseline is the frozen, untrained model (mean $\passat = 0.690$, $\avgq = 0.856$). Notably, the budget $\dbudget$ is enforced only during training via the Lagrange multiplier; at evaluation the policy generates freely, and budget compliance is assessed post-hoc from the resulting question rates.
Checkpoint selection and training curves are in Appendix~\ref{sec:appendix}.
\section{Results}
\label{sec:results}
\begin{figure}[t]
  \centering
  \includegraphics[width=\columnwidth]{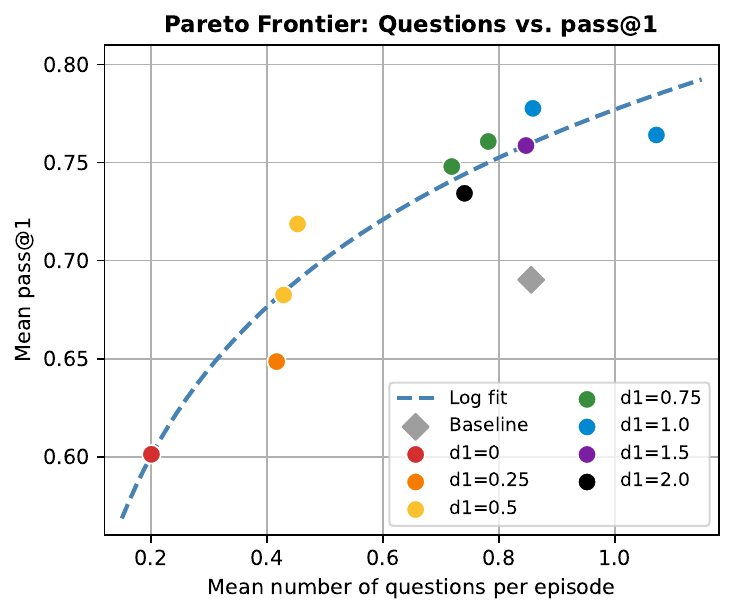}
  \caption{
    Pareto frontier in ($\avgq$, \passat) space for all trained policies and the untrained baseline (gray diamond), evaluated on 417 held-out problems under greedy decoding.
    The log fit illustrates diminishing returns; from $\dbudget{=}1.0$ onward Lagrangian enforcement weakens or ceases.
  }
  \label{fig:pareto}
\end{figure}

\paragraph{Budget compliance and the Pareto frontier.}
Figure~\ref{fig:pareto} plots all trained policies in $(\avgq,\,\passat)$
space.
The key efficiency result is $\dbudget{=}0.5$: at $\avgq = 0.453$
(roughly half the baseline's question rate of $0.856$), two independent runs
achieve mean $\passat \in \{0.683,\, 0.719\}$, neither significantly different from baseline (both $p{\geq}0.10$, paired $t$-test),
confirming that the budget constraint is enforceable at half the question rate
without degrading mean \passat.
Allowing more questions continues to help: $\dbudget{=}0.75$ ($\avgq = 0.782$)
achieves mean $\passat = 0.761$ ($+7.1\%$, best run; other run $+5.8\%$,
both $p{<}0.01$), and $\dbudget{=}1.0$ ($\avgq = 0.859$) achieves
mean $\passat = 0.778$ ($+8.7\%$, best run; other run $+7.4\%$,
both $p{<}0.0001$).
At $\dbudget{=}1.0$, $\lambda_1$ activates at some iterations but does not \textbf{bind} persistently; above $\dbudget{=}1.0$, $\lambda_1 \approx 0$ throughout training (the budget is slack, see Appendix~\ref{sec:appendix_training_curves}), and further budget increases yield no marginal accuracy gain.
In this slack regime, PPO trains without active question-budget pressure yet still improves accuracy over the baseline; these high-budget runs ($\dbudget \in \{1.5, 2.0\}$, $\lambda_1 \approx 0$) therefore serve as an approximate unconstrained-PPO control, indicating that PPO itself contributes to the accuracy gains, while the Lagrangian constraint is what provides budget-targeted control and yields the efficient operating point.
Across all runs, the turn multiplier $\lambda_2$ remained $0$ at every dual update: the maximum batch-mean turn count was $3.33$, below the turn budget $d_2{=}4$. The trained policies are therefore behaviorally question-only, so our main claims rest on the question constraint rather than the turn constraint; $d_2$ serves only as a safeguard against pathological dialogues below the hard cap $T_{\max}{=}6$.

\paragraph{Diminishing returns and the value of clarification.}
The Pareto frontier in Figure~\ref{fig:pareto} follows a log-shaped curve: the marginal \passat gain per additional clarifying question decreases as the budget grows.
The step from $\dbudget{=}0.5$ to $\dbudget{=}0.75$ yields a substantially larger accuracy gain than the step from $\dbudget{=}0.75$ to $\dbudget{=}1.0$, and beyond $\dbudget{=}1.0$ there is no further marginal accuracy gain.
This implies that, beyond a certain budget, the friction cost of each additional clarification outweighs its marginal accuracy benefit.

Among our trained policies, $\dbudget{=}0.75$ is the only one to \textbf{Pareto-dominate} the baseline: it achieves significantly higher \passat ($+5.8$--$+7.1\%$, both $p{<}0.01$) while asking significantly fewer questions ($\avgq \in \{0.719,\,0.782\}$ vs.\ $0.856$; both $p{<}0.01$, paired $t$-test) across both independent runs.
The $\dbudget{=}0.5$ policy achieves comparable \passat at roughly half the baseline question rate (both $p{\geq}0.10$); one run is above baseline and one below, consistent with budget compliance but not a reliable accuracy gain at this budget level.
Appendix~\ref{sec:appendix_episodes} (Episode~2) illustrates Pareto domination concretely.
Policies at $\dbudget \in \{0, 0.25\}$ fall below baseline (details in Appendix~\ref{sec:appendix}), confirming that suppressing clarification too aggressively is counterproductive.

\begin{figure}[h]
  \centering
  \includegraphics[width=\columnwidth]{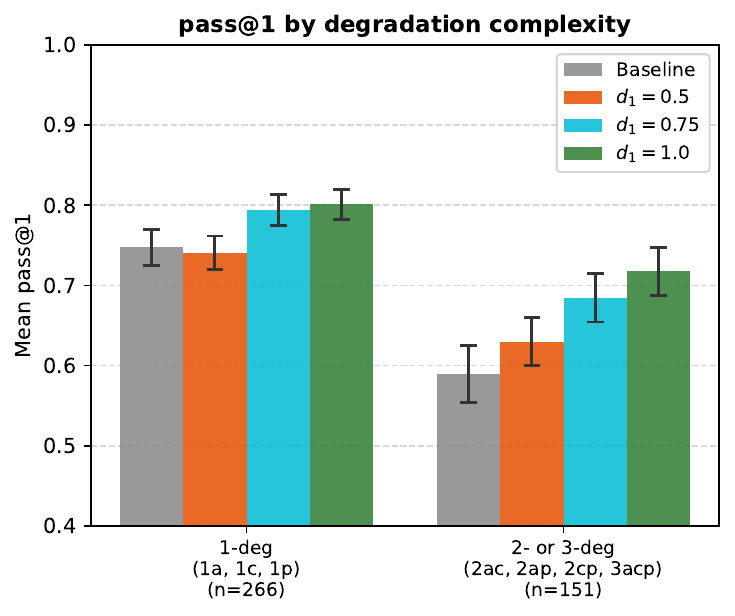}
  \caption{
    Mean \passat by degradation type (417 problems, greedy decoding).
    $\dbudget \in \{0.5, 0.75, 1.0\}$ results averaged across two independent runs.
    Error bars are $\pm 1$ standard error of the mean (SEM).
  }
  \label{fig:deg_group_pass1}
\end{figure}

\paragraph{Performance by degradation complexity.}
Figure~\ref{fig:deg_group_pass1} shows \passat broken down by degradation
complexity group:
\textbf{1-deg} (types 1a, 1c, 1p; $n{=}266$) and \textbf{2- or 3-deg}
(types 2ac, 2ap, 2cp, 3acp; $n{=}151$).
Every policy scores lower on harder multi-degradation problems, where a single
clarifying question can resolve at most one simultaneous spec defect; trained
policies tend to ask more questions on these cases.
Question-count breakdowns by group and by all seven degradation types are in
Appendix~\ref{sec:appendix_deg_type}.

\paragraph{Quality over quantity: better questions and code.}
Controlling for question count (1Q${\to}$1Q episodes, i.e., episodes where both the baseline and the trained policy asked exactly one question), $\dbudget{=}0.75$ achieves
$+7.1\%$ over baseline ($p{<}0.001$, $n{=}490$ episodes pooled across both runs, paired $t$-test), showing
gains do not stem from higher question counts.
The per-episode breakdown (Appendix~\ref{sec:appendix_deg_type}) is consistent with better question targeting and code quality under ambiguity.

\section{Conclusion}
\label{sec:conclusion}

CMDP post-training with PPO-Lagrangian produces a Pareto frontier in $(\avgq, \passat)$ space: trained policies simultaneously achieve higher accuracy and lower question rates than the untuned baseline.
Two structural findings extend this result.
First, policies where the Lagrangian is actively enforced ($\dbudget \in \{0.5,\, 0.75\}$, $\lambda_1 {>} 0$) ask
more questions on harder multi-degradation problems while asking fewer questions
overall than the baseline (Figure~\ref{fig:deg_group_pass1} and Appendix~\ref{sec:appendix_deg_type}),
suggesting that CMDP training reallocates question budget toward more
informative contexts rather than reducing it.
Second, gains are driven by question quality and code generation ability rather than question count:
trained policies significantly outperform the baseline even when both ask exactly one question.

The Pareto-dominant operating point ($\dbudget{=}0.75$, within the binding regime where $\lambda_1{>}0$) demonstrates that active Lagrangian enforcement achieves simultaneous quality and efficiency gains.
The log-shaped Pareto frontier further suggests a practical design principle: there is a budget threshold beyond which the friction cost of additional clarifications outweighs their marginal accuracy benefit, and operating beyond it yields no gain.
Future work should examine this framework on naturally occurring ambiguity,
larger model scales, and human rather than simulator oracles.
A natural extension is a \emph{dynamic} budget that adapts to the estimated complexity or ambiguity of each problem, rather than enforcing a fixed expected question count across all episodes.

\section*{Limitations}

\paragraph{Run-to-run variance.}
Independent training runs at the same budget ($\dbudget \in \{0.5, 0.75,
1.0\}$) converge to notably different mean \passat values (e.g., $0.683$ vs.\
$0.719$ at $\dbudget{=}0.5$; $0.778$ vs.\ $0.764$ at $\dbudget{=}1.0$),
indicating sensitivity to random initialisation.
This limits the reliability of conclusions drawn from any single run and
motivates future work on more stable training procedures.

\paragraph{Training scale and instability.}
All policies were trained under significant compute constraints: only 80
iterations with a mini-batch size of 8 episodes and 32 episodes per
rollout iteration.
Small batches produce high-variance gradient estimates, which destabilises
the dual-ascent updates for the Lagrange multipliers and likely contributes
to the run-to-run variance noted above.
Larger batch sizes, more iterations, and learning-rate schedules tuned for
the Lagrangian dynamics would likely improve both convergence reliability
and final performance.

\paragraph{LLM-based question simulator.}
The oracle that answers clarifying questions is GPT-4o-mini running at
temperature~0, making its responses deterministic given the question and
ground-truth specification.
This eliminates answer inconsistency as a noise source, but a greedy
deterministic simulator may not reflect the variability of real human
responses: a human might paraphrase, hedge, or volunteer adjacent information
differently across interactions, all of which would affect the policy's
learned question-asking strategy.

\paragraph{Single benchmark and small training set.}
All results are on HumanEvalComm, a single code-generation benchmark with a
specific set of synthetic degradation types.
Whether the learned strategies and efficiency gains generalise to other task
types or to naturally occurring ambiguity is untested.
The training pool is also small: 303 degraded instances from 64 base problems,
smaller than the 417-problem held-out test set.
Whether this scale is sufficient for reliable convergence is unclear, and likely
contributes to the run-to-run variance observed across seeds.

\paragraph{Scale.}
Experiments use a 7B-parameter model with LoRA fine-tuning.
Whether the CMDP approach scales to larger models, whether the emergent
budget-allocation behaviour persists, and whether the Lagrange multiplier
dynamics remain stable at larger scale are open questions.

\paragraph{Discrete budget coverage.}
Due to compute constraints, we train at only seven discrete budget values
($\dbudget \in \{0, 0.25, 0.5, 0.75, 1.0, 1.5, 2.0\}$); the log fit in
Figure~\ref{fig:pareto} interpolates between them.
The true shape of the Pareto frontier between trained budgets is unverified;
the optimal trade-off curve could differ meaningfully from the fitted log curve
in these intervals.

\paragraph{Disentangling question quality from code quality.}
PPO training jointly optimises question-asking and code generation, making it
difficult to attribute gains solely to improved clarification behaviour.
Cases where a policy asks comparable questions to the baseline yet produces
better code suggest that general coding ability also improves under PPO.
Future work should design controlled ablations, for instance freezing the
code-generation head and training only the question-asking policy, to cleanly
separate these two sources of improvement.

\section*{Acknowledgments}

The authors used Claude (Anthropic) for assistance with writing and editing during this project. All experimental design, analysis, implementation, math, and final verification were performed by the authors.

\bibliography{custom}

\appendix

\section*{Appendix}

\section{Implementation Details}
\label{sec:appendix}

Table~\ref{tab:hyperparams} consolidates all hyperparameters used across all
training runs.
All policies use identical hyperparameters; only the budget $\dbudget$ varies.

\paragraph{CMDP formal definitions.}
\textbf{State.}
At turn $t$, $s_t$ is the full conversation history: the degraded problem specification $x$
together with all prior question-answer pairs $\{(q_1,u_1),\ldots,(q_{t-1},u_{t-1})\}$.
\textbf{Transition function.}
For \texttt{[ANSWER]} actions, the episode terminates deterministically.
For \texttt{[ASK]} actions, $s_{t+1}$ incorporates the GPT-4o-mini user simulator's
response to $q_t$; the simulator runs at temperature $= 0$, making its responses
deterministic given the question and ground-truth specification.
\textbf{Reward.}
$R(s_t, a_t) = \rho(c_t)$ if $a_t = \texttt{[ANSWER]}\;c_t$, else $0$,
where $\rho(c_t) \in [0,1]$ is \passat: the fraction of hidden test assertions passed.
Reward is non-zero only at the terminal step.
The training signal optimizes \passat.

\paragraph{Policy and value heads.}
We parameterize the policy as Qwen2.5-Coder-7B-Instruct~\citep{hui2024qwen25codertechnicalreport}
with LoRA~\citep{hu2021loralowrankadaptationlarge} ($r = 16$, $\alpha = 32$) applied to all seven attention
and MLP projection layers (${\approx}40$M of 7B parameters trainable).
Three value heads $V^R$, $V^{C_q}$, $V^{C_t}$, each implemented as
$\mathrm{Linear}(h, 1024) \to \mathrm{GELU} \to \mathrm{LayerNorm}(1024)
\to \mathrm{Linear}(1024, 1)$ over the final-layer last-prompt-token hidden state,
estimate expected future returns for the reward, question-cost, and turn-cost streams.
Output layers are zero-initialized; value MSE gradients backpropagate through the
shared LoRA representation.
A frozen reference copy (base weights, no LoRA) runs on a second GPU for KL
regularization; a rollout copy (LoRA weights synced each iteration) handles episode
collection.
Both $\pi_\theta$ and $\pi_{\theta_{\mathrm{old}}}$ are computed via clean forward
passes, bypassing HuggingFace logit processors that would otherwise corrupt
log-probability scales.

\paragraph{PPO surrogate and prefix exclusion.}
The policy is updated via the clipped PPO objective~\citep{schulman2017proximalpolicyoptimizationalgorithms}
($\varepsilon = 0.2$):
\begin{equation}
\begin{aligned}
L^{\mathrm{PPO}}
= -\mathbb{E}_t\!\Bigl[
\min\Bigl(
&r_t \hat{A}^{\mathrm{lag}}_t, \\
&\operatorname{clip}(r_t,\,
1{-}\varepsilon,\,
1{+}\varepsilon)\,
\hat{A}^{\mathrm{lag}}_t
\Bigr)
\Bigr].
\end{aligned}
\label{eq:ppo_loss}
\end{equation}
where
\[
r_t(\theta)
=
\frac{
\pi_\theta(a_{\mathrm{cont},t}\mid s_t,\mathrm{pref}_t)
}{
\pi_{\theta_{\mathrm{old}}}(a_{\mathrm{cont},t}\mid s_t,\mathrm{pref}_t)
}
\]
is the probability ratio over \emph{continuation tokens only}; the forced
\texttt{[ASK]}/\texttt{[ANSWER]} prefix is excluded from both numerator and
denominator.

\begin{table}[t]
  \centering
  \footnotesize
  \setlength{\tabcolsep}{3pt}
  \begin{tabularx}{\columnwidth}{@{}l l >{\raggedleft\arraybackslash}X@{}}
  \toprule
  \textbf{Component} & \textbf{Hyperparameter} & \textbf{Value} \\
  \midrule
  \multirow{5}{*}{Policy (LoRA)}
    & Rank $r$                    & 16 \\
    & Scale $\alpha$              & 32 \\
    & Trainable parameters        & $\approx$40M \\
    & Learning rate               & $5 \times 10^{-6}$ \\
    & LR warmup                   & 20 steps (linear) \\
  \midrule
  \multirow{4}{*}{Value heads}
    & Hidden dim                  & 1024 \\
    & Activation                  & GELU \\
    & Learning rate               & $10^{-4}$ \\
    & Output layer init           & zeros \\
  \midrule
  \multirow{7}{*}{PPO}
    & Clip $\varepsilon$          & 0.2 \\
    & Mini-batch size             & 8 \\
    & Max epochs / iteration      & 4 \\
    & KL early-stop threshold     & 0.05 \\
    & $\beta_{\mathrm{KL}}$       & 0.25 \\
    & $\beta_H$                   & 0.01 \\
    & Value loss coefficient      & 0.5 \\
  \midrule
  \multirow{2}{*}{GAE}
    & $\gamma$                    & 1.0 \\
    & $\lambda_{\mathrm{GAE}}$    & 0.95 \\
  \midrule
  \multirow{5}{*}{Lagrangian}
    & $\eta_{\lambda_1}$          & 0.1 \\
    & $\eta_{\lambda_2}$          & 0.005 \\
    & $\lambda_{1,\max}$          & 10 \\
    & $\lambda_{2,\max}$          & 5 \\
    & Turn budget $d_2$           & 4 \\
  \midrule
  \multirow{5}{*}{Training}
    & Episodes / iteration        & 32 \\
    & Total iterations            & 80 \\
    & Checkpoint frequency        & every 5 iterations \\
    & Rollout temperature         & 0.8 \\
    & Eval temperature            & 0 (greedy) \\
  \midrule
  \multirow{2}{*}{Episode}
    & Max turns $T_{\max}$        & 6 \\
    & Optimizer                   & 8-bit AdamW \\
  \midrule
  Gradient clipping & $\ell_2$ norm & 1.0 \\
  \bottomrule
  \end{tabularx}
  \caption{Hyperparameters used for all training runs.}
  \label{tab:hyperparams}
  \end{table}

\paragraph{Routing gradient.}
Routing and generation share the same LoRA weights: the prefix scores used to select
\texttt{[ASK]} vs.\ \texttt{[ANSWER]} at rollout time are computed from the same
attention and MLP projections that PPO updates via continuation-token gradients.
When $\lambda_1$ rises, the $-\lambda_1\tilde{A}^{C_q}_t$ term penalises
\texttt{[ASK]}-turn continuations; the resulting weight updates shift the model's
logits at the routing step in the next iteration.
Budget enforcement thus propagates to routing through shared parameters across
training iterations, which is the standard mechanism by which PPO shapes
structured generation behaviour.
As a concrete two-token example, suppose the policy emits \texttt{[ASK]~what type?}.
The \texttt{[ASK]} prefix tokens are excluded from the ratio $r_t$, so
the gradient never directly rewards or penalises the \emph{decision} to ask; only
the continuation \texttt{what type?} enters the loss.
Yet because the prefix logits and the continuation are produced by the same shared
LoRA weights, updating those weights to make a useful \texttt{what type?} more likely
also perturbs the logits that produced \texttt{[ASK]}: the routing decision is
coupled to the continuation gradient and is shaped as a side effect of it, rather
than being fixed.
The routing decision is therefore learned indirectly, through the shared parameters,
rather than optimised directly against the sparse \passat signal, which would
otherwise attribute the entire episode's outcome to a single decision token.

\paragraph{Episode protocol.}
Each episode begins with the agent receiving a degraded problem specification.
At each turn:
(1) the agent generates an action that begins with \texttt{[ASK]} or
\texttt{[ANSWER]}, selected via prefix-scored routing;
(2) \textbf{if \texttt{[ASK]}}, the question is passed to the GPT-4o-mini user
simulator, which holds the complete specification and returns a factual answer
together with an atomic question count $N$, setting $C_q = N$;
(3) \textbf{if \texttt{[ANSWER]}}, code is executed against hidden tests, yielding
$\rho = k/n$ ($k$ of $n$ assertions passing), and the episode terminates.
Episodes reaching $T_{\max} = 6$ terminate with $R = 0$.

\paragraph{Atomic question counting.}
The simulator appends \texttt{QUESTION\_COUNT:~N} to its reply, where $N$ is the
number of distinct information requests in the \texttt{[ASK]} turn.
Multi-question turns are charged $C_q = N > 1$, preventing budget exploitation by
bundling questions.

\paragraph{Prefix-scored routing.}
Given $s_t$, the model performs a forward pass over the prompt and scores each prefix
via log-softmax over the final-position logits.
Under the Qwen2.5-Coder tokenizer, \texttt{[ASK]~} and \texttt{[ANSWER]~} each
tokenize as four tokens but share the first (\texttt{[}), so the model performs a
second forward step via KV cache to score the second distinguishing token of each
prefix (\texttt{ASK} vs.\ \texttt{ANSWER}).
The prefix with the higher two-token sum of log-softmax values is selected; this is a
heuristic approximation to the full four-token joint log-probability and is distinct
from hard-constraint decoding algorithms that guarantee structural constraints
(tries, finite-state acceptors, grammar-constrained decoding).
At evaluation (temperature $= 0$), the higher-scoring prefix is selected greedily;
during training rollouts (temperature $= 0.8$), the prefix is sampled from a
Boltzmann distribution over the two scores.
Ties at greedy decoding default to \texttt{[ASK]}.
The model generates freely after the chosen prefix.

\paragraph{GAE recurrence~\citep{schulman2016highdimensionalcontinuous}.}
Three advantage streams are computed via GAE for $X \in \{R, C_q, C_t\}$:
\begin{equation}
\begin{aligned}
\delta^X_t
&= x_t + \gamma V^X(s_{t+1}) - V^X(s_t), \\
A^X_t
&= \sum_{l \geq 0}
(\gamma\lambda_{\mathrm{GAE}})^l\,\delta^X_{t+l}.
\end{aligned}
\label{eq:gae}
\end{equation}
where $x_t$ is the stream-specific signal at turn $t$: $R(s_t,a_t)$ for $X{=}R$, $C_q(a_t)$ for $X{=}C_q$, and $C_t(a_t)$ for $X{=}C_t$.
The $\gamma{=}1.0$, $\lambda_{\mathrm{GAE}}{=}0.95$ setting introduces an effective per-turn discount of $0.95^\tau$ on advantages at turn depth $\tau$ from the episode end.

\paragraph{Double whitening.}
Let $\hat{A}^{\mathrm{lag}}_t = \mathrm{whiten}(A^{\mathrm{lag}}_t)$ denote the result of whitening $A^{\mathrm{lag}}_t$ once more before the PPO surrogate for numerical stability; this second pass scales all terms by $1/\sqrt{1+\lambda_1^2+\lambda_2^2}$ (under stream independence), so $\lambda_1$ shifts the advantage threshold rather than amplifying the absolute gradient magnitude of the constraint term.

\paragraph{Full training objective.}
\begin{equation}
L(\theta) = L^{\mathrm{PPO}} + \beta_{\mathrm{KL}} L_{\mathrm{KL}} - \beta_H L_H
  + 0.5\,L^{\mathrm{vf}},
\label{eq:total_loss}
\end{equation}
where $L_{\mathrm{KL}} = \max(0, \mathbb{E}[\log\pi_\theta - \log\pi_{\mathrm{ref}}])$
is a sequence-level KL from the frozen reference (clamped non-negative,
$\beta_{\mathrm{KL}} = 0.25$); $L_H = -\mathbb{E}[\log\pi_\theta / |a_{\mathrm{cont}}|]$
a length-normalized entropy bonus ($\beta_H = 0.01$); and $L^{\mathrm{vf}}$ total
value MSE across all three heads.
All terms are over continuation tokens only.
Approx-KL is $\hat{D}_{\mathrm{KL}} = \mathbb{E}[(r_t - 1) - \log r_t]$; if
$\hat{D}_{\mathrm{KL}} > 0.05$ per mini-batch, remaining epochs for that iteration
are skipped.
Gradient $\ell_2$ norm is clipped to 1.0.

\paragraph{Dual update mechanics.}
The $\lambda$ values entering $\hat{A}^{\mathrm{lag}}$ (via Eq.~\eqref{eq:lag_adv}) within the
PPO update are therefore the post-dual-update multipliers of the current iteration.
When $\avgq > \dbudget$, $\lambda_1$ rises, shifting the sign threshold in
$\hat{A}^{\mathrm{lag}}$: some \texttt{[ASK]} continuations that ranked as positive flip to
negative, making them less likely to be reinforced by PPO.
The absolute gradient scale is fixed by the second whitening and is not affected
by $\lambda_1$.

\paragraph{Per-iteration training steps.}
Each PPO-Lagrangian iteration proceeds as:
(1)~sync LoRA weights to the rollout model;
(2)~sample a batch of 32 episodes from the training pool (user-simulator API calls concurrent via \texttt{asyncio});
(3)~compute GAE for all three streams;
(4)~update $\lambda_1$, $\lambda_2$ via dual ascent (Eqs.~\eqref{eq:dual1}--\eqref{eq:dual2});
(5)~run up to 4 PPO epochs with early-exit on approx-KL;
(6)~step the LR scheduler.
Policy and value heads use 8-bit AdamW~\citep{dettmers20228bitoptimizersblockwisequantization} (LR $5{\times}10^{-6}$ and $10^{-4}$; 20-step linear warmup for the policy, constant thereafter).

\paragraph{Train/eval split.}
All degraded variants of the same base HumanEval problem are kept in the same
split, preventing the model from seeing the ground-truth specification of an
eval problem in a different degraded form during training.
A fixed set of 52 validation problems (stratified by degradation type,
sampled with a fixed seed) is used for all policies, leaving 417 held-out
problems for the main eval results reported here.

\paragraph{Training setup.}
Training uses 303 degraded instances from the 64 base problems not held out for evaluation (not all degradation types are available for every base problem; rare types such as \texttt{3acp} exist for only 12 of 164 base problems), running 80 PPO-Lagrangian iterations per budget (${\approx}11$h on 2$\times$A100-40GB; hyperparameters in Table~\ref{tab:hyperparams}).
The reward signal is exclusively \passat.

\paragraph{Checkpoint selection.}
Checkpoints are saved every 5 iterations; the 52-problem validation set (stratified
by degradation type, seed 99) is evaluated under greedy decoding at every 10th
checkpoint (iter 9, 19, $\ldots$, 79).
A checkpoint \emph{satisfies} the budget if $\avgq^{\mathrm{val}} \leq \dbudget$ on
the val set under greedy decoding; the primary criterion selects the highest val
pass@1 among satisfying checkpoints.
If no satisfying checkpoint exists (as at $\dbudget = 0$), the fallback selects the
checkpoint with the lowest $\avgq^{\mathrm{val}}$; the $\dbudget = 0$ result
therefore reflects training \emph{toward} zero clarification, not a converged
zero-question policy.
The $\dbudget = 0.25$ run is excluded from the frontier: it satisfied the budget at validation but violated it at test time ($\avgq^{\mathrm{test}} = 0.417 > 0.25$).

\paragraph{Pareto frontier and evaluation.}
All seven budget levels are studied;
all runs use identical hyperparameters, varying only $\dbudget$.
All policies are evaluated on the 417-problem held-out test set
(Section~\ref{sec:setup}); statistical comparisons use paired $t$-tests per run.
The baseline is untuned Qwen2.5-Coder-7B-Instruct under identical multi-turn
conditions (prefix-scored routing, $T_{\max} = 6$, greedy decoding).
Budget compliance is assessed per run on the test set via one-sample $t$-test
($H_0{:}\;\mu_q \leq \dbudget$, $\alpha{=}0.05$).
At $\dbudget{=}0.75$, neither run significantly exceeds the budget
(run~1: $\avgq{=}0.782$, $p{=}0.15$; run~2: $\avgq{=}0.719$, $p{=}0.86$);
both are statistically budget-compliant ($n{=}417$, SE~${\approx}0.03$).
At $\dbudget{=}1.0$, run~1 satisfies the budget ($\avgq{=}0.859$, well below budget),
while run~2 significantly exceeds it ($\avgq{=}1.072$, $p{=}0.027$).
The $\dbudget = 0$ policy falls below the baseline and is dominated by $\dbudget = 0.5$;
it does not Pareto-dominate the baseline.
The reported frontier ($\dbudget \in \{0.5,\, 0.75,\, 1.0,\, 1.5,\, 2.0\}$) is anchored
to the budget-satisfying run at each level; gains from budget-violating runs are
reported separately for completeness.

\paragraph{$\dbudget = 0.25$ run details.}
The single run reached $\avgq^{\mathrm{val}} = 0.25$ at iter\_0079 (val pass@1 $= 0.481$, budget-feasible at val selection), but test-time $\avgq = 0.417 > 0.25$; it is excluded from the Pareto frontier because $\avgq^{\mathrm{test}} > \dbudget$.

\section{Training Curves}
\label{sec:appendix_training_curves}

\begin{figure*}[p]
  \centering
  \includegraphics[width=\textwidth]{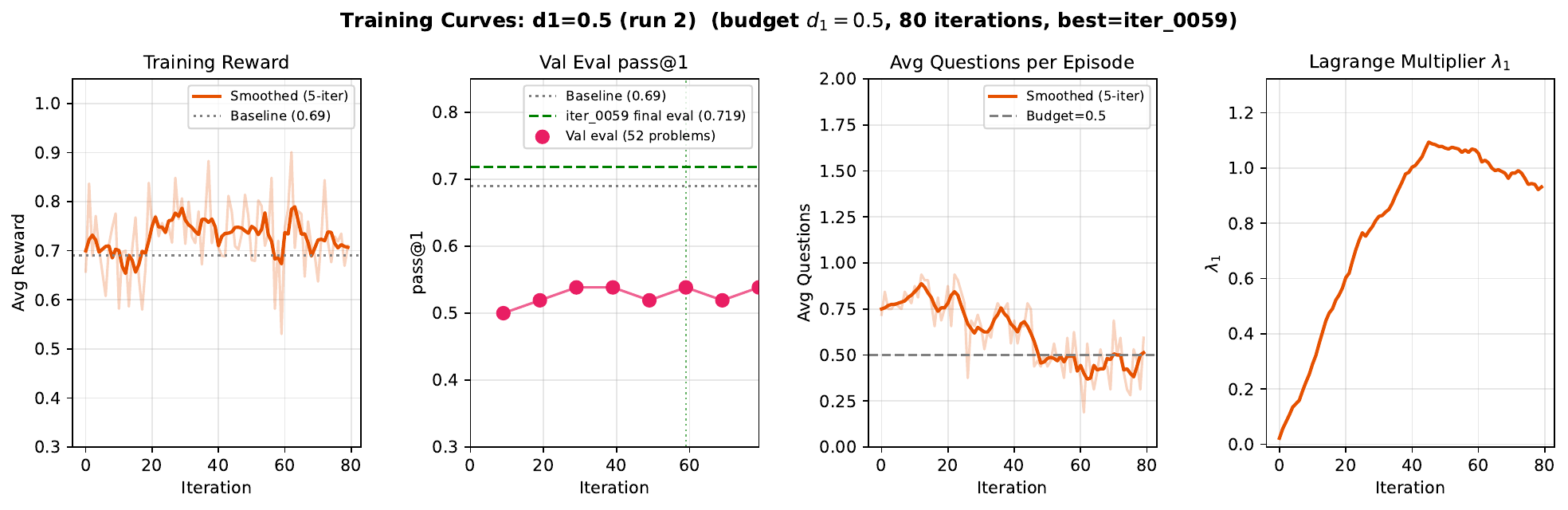}
  \caption{Training curves for $\dbudget = 0.5$.}
  \label{fig:tc_d05}
\end{figure*}

\begin{figure*}[t]
  \centering
  \includegraphics[width=\textwidth]{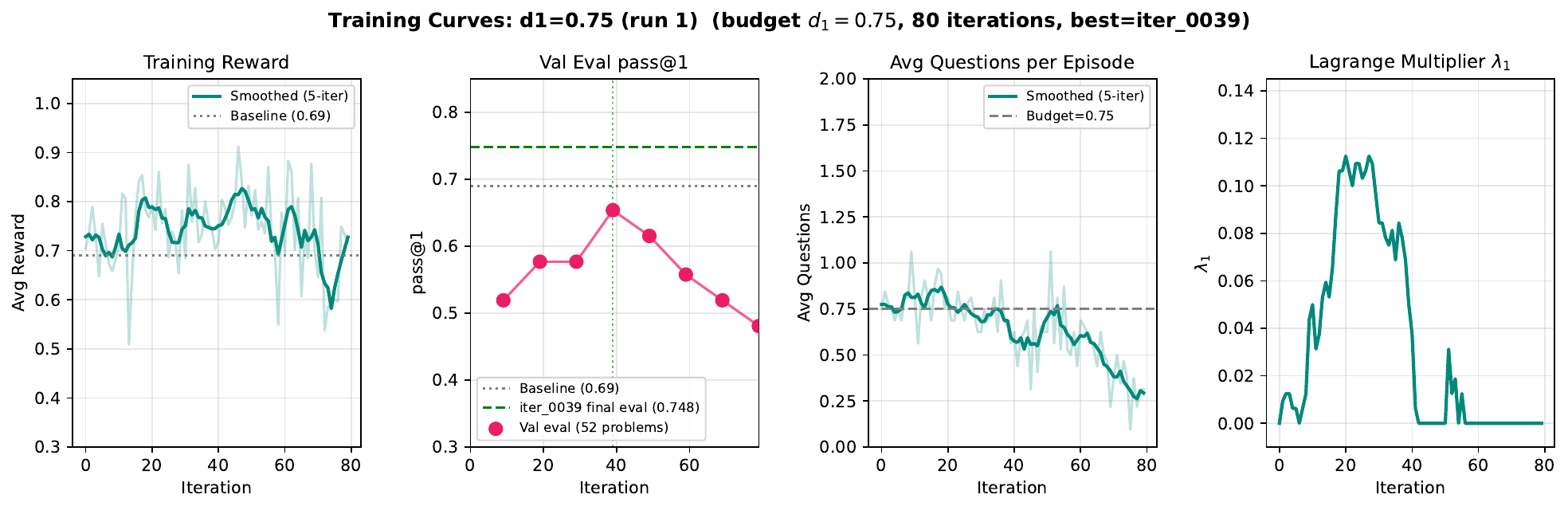}
  \caption{Training curves for $\dbudget = 0.75$.}
  \label{fig:tc_d075}
\end{figure*}

\begin{figure*}[t]
  \centering
  \includegraphics[width=\textwidth]{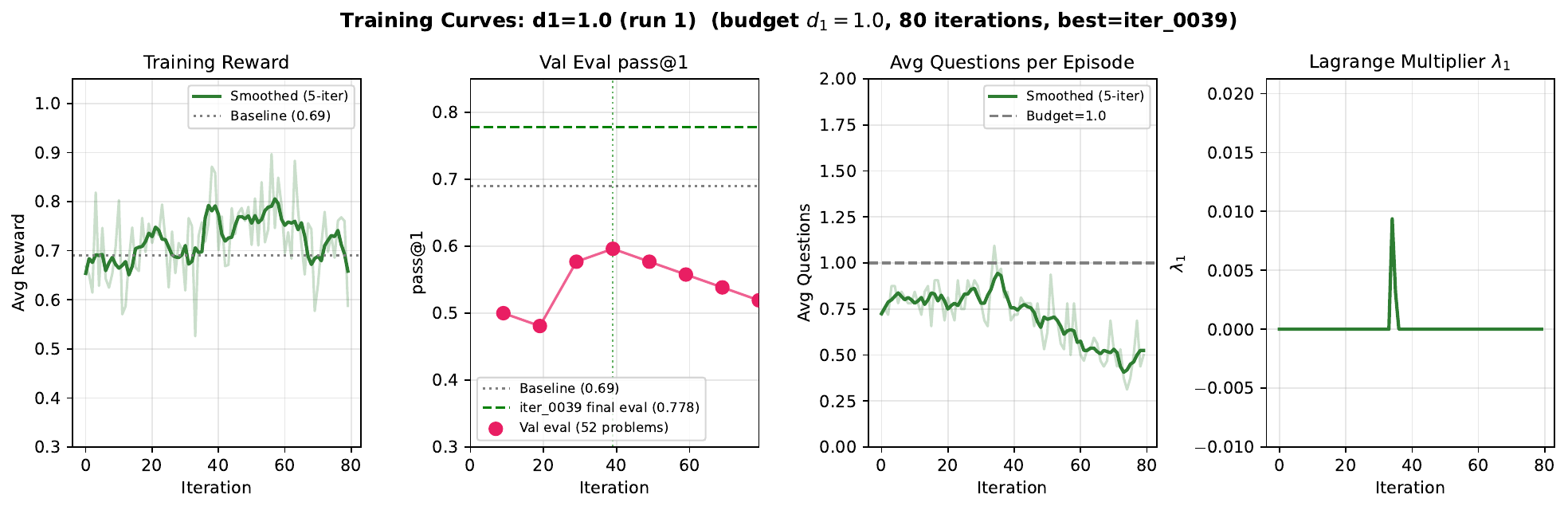}
  \caption{Training curves for $\dbudget = 1.0$.}
  \label{fig:tc_d10}
\end{figure*}

Figures~\ref{fig:tc_d05}--\ref{fig:tc_d10} show training dynamics for
representative policies at $\dbudget \in \{0.5, 0.75, 1.0\}$.
These three budgets span the efficient region of the frontier: $\dbudget{=}0$ never achieves budget compliance within 80 iterations and the single $\dbudget{=}0.25$ run violates the budget at test time,
budgets strictly above $\dbudget = 1.0$ show $\lambda_1 = 0$ throughout training
(the constraint never binds), and $\{0.5, 0.75, 1.0\}$ are the only budgets
trained with two independent runs.
Each figure shows four panels across the 80 training iterations:
(1)~mean episode reward,
(2)~mid-training validation \passat evaluated every 10 iterations with the
selected checkpoint marked,
(3)~mean questions per episode $\avgq$ with the budget $\dbudget$ as a dashed
reference line, and
(4)~the Lagrange multiplier $\lambda_1$.

The $\lambda_1$ panels directly confirm the budget compliance story from
Section~\ref{sec:results}: for $\dbudget = 1.0$, $\lambda_1$ activates at some iterations but does not persist throughout training,
while for $\dbudget \in \{0.5, 0.75\}$, $\lambda_1$ rises early and stabilises at a persistently positive value,
reflecting the dual-ascent feedback loop enforcing the budget.

\section{Degradation Type Breakdown}
\label{sec:appendix_deg_type}

At $\dbudget{=}1.0$, mean \passat rises on zero-question episodes ($73.7\%$ baseline
${\to}$ $81.1\%$) and on one-question episodes ($67.8\%$ ${\to}$ $77.4\%$),
consistent with the model learning to write better code under ambiguity and to
formulate more targeted questions when clarification is warranted.

\begin{figure*}[p]
  \centering
  \includegraphics[width=\textwidth]{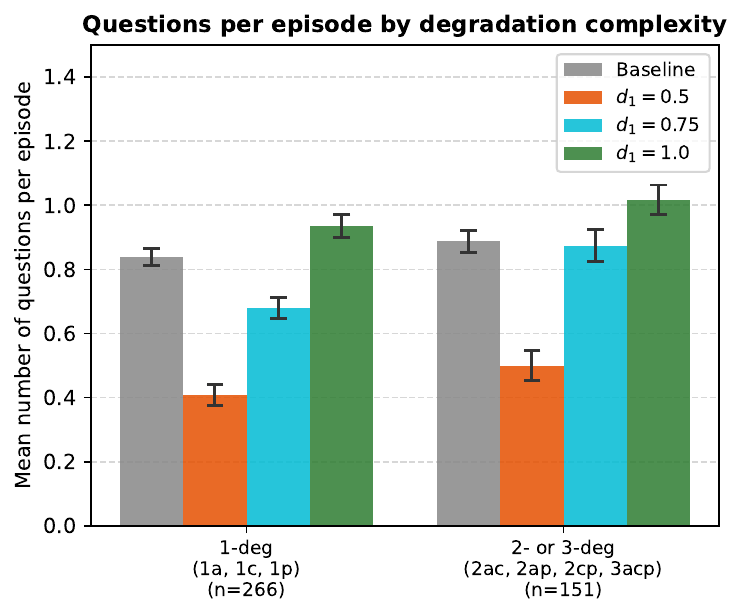}
  \caption{
    Mean number of questions per episode by degradation complexity group,
    evaluated on the 417-problem held-out set under greedy decoding.
    Settings match Figure~\ref{fig:deg_group_pass1}.
  }
  \label{fig:deg_group_qs}
\end{figure*}

\begin{figure*}[!t]
  \centering
  \includegraphics[width=0.9\textwidth]{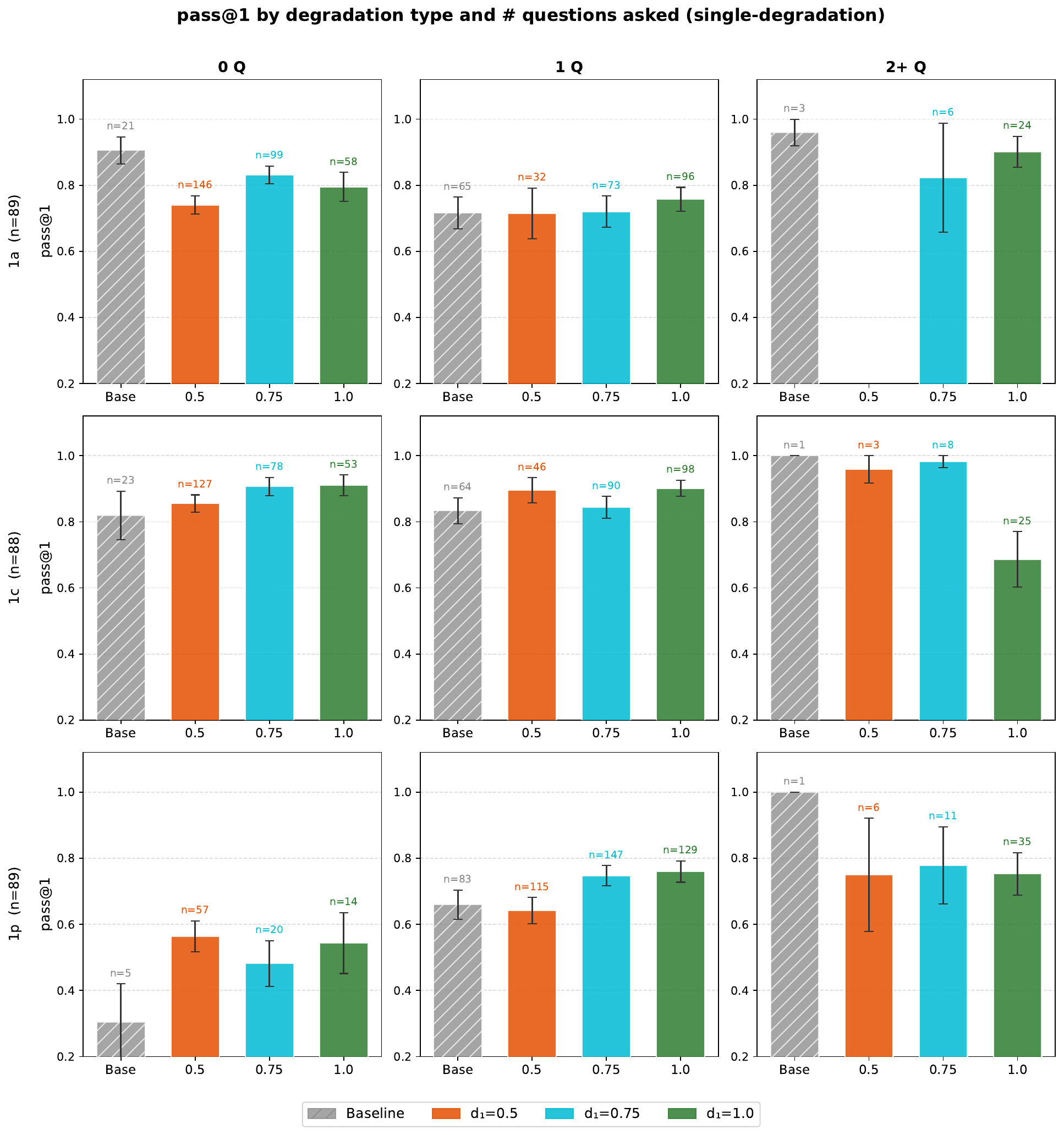}
  \caption{
    \passat by degradation type and number of questions asked,
    for single-degradation types (1a, 1c, 1p).
    Columns correspond to episodes where 0, exactly~1, or 2+
    clarifying questions were asked.
    Per-run pooling: for policies with two independent runs, each run
    contributes its episodes independently, classified by that run's question
    count, then pooled before computing means.
    Error bars are $\pm 1$ SEM.
  }
  \label{fig:deg_type_by_qcount_1deg}
\end{figure*}

\begin{figure*}[!t]
  \centering
  \includegraphics[width=0.9\textwidth]{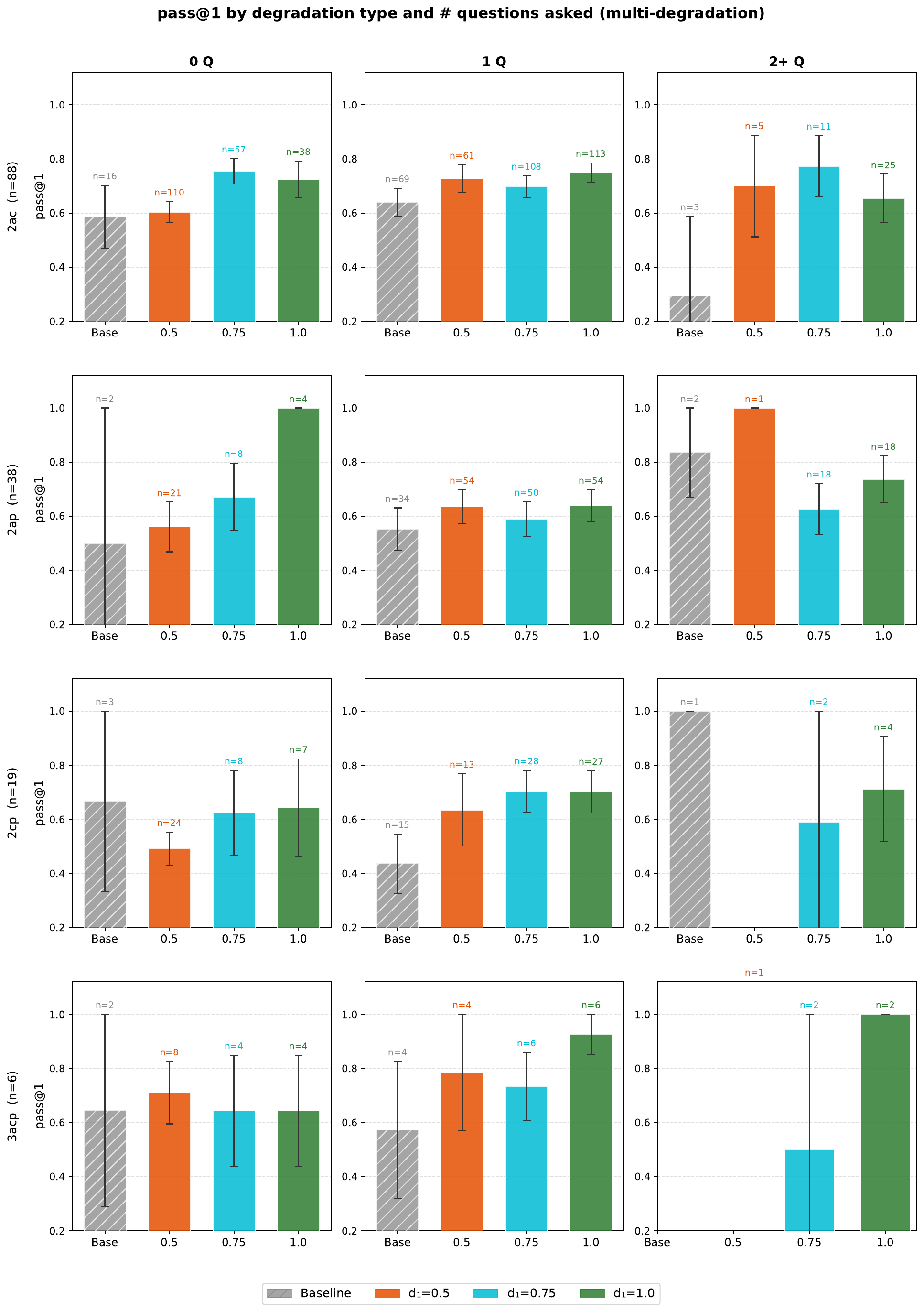}
  \caption{
    \passat by degradation type and number of questions asked,
    for multi-degradation types (2ac, 2ap, 2cp, 3acp).
    Same layout and pooling as Figure~\ref{fig:deg_type_by_qcount_1deg}.
    Cells with very small~$n$ (notably 3acp and the 2+~Q column) should
    be interpreted cautiously.
  }
  \label{fig:deg_type_by_qcount_multideg}
\end{figure*}

Figures~\ref{fig:deg_type_by_qcount_1deg} and~\ref{fig:deg_type_by_qcount_multideg} show \passat broken down by all seven
individual degradation types (rows) and question count: 0~questions,
exactly~1~question, and 2+~questions (columns).
For each policy, episodes are partitioned per run by that run's question count
and pooled across both runs before computing means.
This finer-grained view reveals heterogeneous behaviour within the 1-deg group
that is obscured in the aggregate Figures~\ref{fig:deg_group_pass1} and~\ref{fig:deg_group_qs}:
ambiguity problems (1a) are near-ceiling for all models regardless of question
count; contradiction problems (1c) show consistent gains for trained policies
in both the 0Q and 1+Q columns, reflecting improved code-generation quality
for contradictory specs; and truncation problems (1p) show that the 0Q
performance gap between trained policies and the baseline is largely a
selection effect i.e. trained policies have learned to skip asking on the
easier truncation cases, while the baseline skips asking at random.

\FloatBarrier

\FloatBarrier
\clearpage
\section{Representative Episode Comparisons}
\label{sec:appendix_episodes}

Each episode below shows the degraded specification presented to the agent,
followed by the conversation.
All episodes are from the 417-problem held-out evaluation set under greedy
decoding (temperature~$=0$).

\paragraph{Episode 1. Interaction format; targeted vs.\ edge-case questioning
(HumanEval/133/prompt1p, type: 1p).}
The spec body is truncated mid-sentence; the task description is entirely
missing.

\begin{specbox}
\begin{lstlisting}
def candidate(lst):
    """You are given a list of numbers.
    You need to return the sum of squared
    """
\end{lstlisting}
\end{specbox}

\noindent\textbf{Baseline} (pass@1~$=0.00$, 1 question):

\begin{askbox}{Turn 1 [ASK]}
\begin{lstlisting}
What should be the behavior if the input list is empty?
Should it return 0 or raise an error?
\end{lstlisting}
\end{askbox}
\begin{simbox}
\begin{lstlisting}
The behavior for an empty input list is not specified
in the provided documentation.
\end{lstlisting}
\end{simbox}
\begin{answerbox}{Turn 2 [ANSWER]}
\begin{lstlisting}
0
\end{lstlisting}
\end{answerbox}

\noindent\textbf{Trained policy $\dbudget{=}1.0$} (pass@1~$=1.00$, 1 question):

\begin{askbox}{Turn 1 [ASK]}
\begin{lstlisting}
What is the task? Specifically, what should be done
with the list of numbers?
\end{lstlisting}
\end{askbox}
\begin{simbox}
\begin{lstlisting}
The task is to return the sum of the squared numbers
in the given list after rounding each element to the
upper integer (ceiling) first.
\end{lstlisting}
\end{simbox}
\begin{answerbox}{Turn 2 [ANSWER]}
\begin{lstlisting}
import math
def candidate(lst):
    return sum(math.ceil(x) ** 2 for x in lst)
\end{lstlisting}
\end{answerbox}

The baseline does not yet know what the function is supposed to compute,
yet asks about an edge case, a question that is only meaningful once the
core task is understood.
It receives an uninformative answer and submits a literal~\texttt{0}.
The trained policy identifies that the \emph{entire task description}
is missing and recovers it in one targeted question.

\paragraph{Episode 2. Pareto efficiency: fewer questions, higher accuracy
(HumanEval/45/prompt1p, type: 1p).}
The spec description is truncated to a single ambiguous word: the docstring
reads only \texttt{``Given length of a side and high, return.''}, omitting the
return type and leaving \texttt{high} undefined.
The baseline spends its question budget confirming the obvious (\texttt{high}
$=$ height), but its answer still fails; the $\dbudget{=}0.5$ policy answers
immediately and correctly.

\begin{specbox}
\begin{lstlisting}
def triangle_area(a, h):
    """Given length of a side and high, return.
    """
\end{lstlisting}
\end{specbox}

\noindent\textbf{Baseline} (pass@1~$=0.00$, 1 question):

\begin{askbox}{Turn 1 [ASK]}
\begin{lstlisting}
Is the high meant to be the height of the triangle?
\end{lstlisting}
\end{askbox}
\begin{simbox}
\begin{lstlisting}
Yes, in this context, 'high' refers to the height of the triangle.
\end{lstlisting}
\end{simbox}
\begin{answerbox}{Turn 2 [ANSWER]}
\begin{lstlisting}
0.5 * a * h
\end{lstlisting}
\end{answerbox}

\medskip
\noindent\textbf{Trained policy $\dbudget{=}0.5$} (pass@1~$=1.00$, 0 questions):

\begin{answerbox}{Turn 1 [ANSWER]}
\begin{lstlisting}
def triangle_area(a, h):
    return 0.5 * a * h
\end{lstlisting}
\end{answerbox}

The baseline wastes its question budget confirming that \texttt{high} means
height, information that does not change the solution, and then outputs a
bare expression instead of a function definition, scoring 0.
The trained policy infers the formula directly from the signature and returns
a complete, correct function without asking anything.
This is one concrete instance of the Pareto dominance shown in
Figure~\ref{fig:pareto}.

\paragraph{Episode 3. Multi-degradation problems require more questions
(HumanEval/126/prompt2ap, type: 2ap).}
The spec has two simultaneous degradations: the sort direction is ambiguous
\emph{and} the duplicate-handling rule is truncated.
A single question can resolve at most one.

\begin{specbox}
\begin{lstlisting}
def candidate(lst):
    '''
    Given a list of numbers, return whether or not they are
    sorted in a specific order. If list has certain duplicate
    of the same number,
    '''
\end{lstlisting}
\end{specbox}

\noindent\textbf{Baseline} (pass@1~$=0.85$, 1 question):

\begin{askbox}{Turn 1 [ASK]}
\begin{lstlisting}
What is the specific order that the list should be sorted in?
\end{lstlisting}
\end{askbox}
\begin{simbox}
\begin{lstlisting}
The list should be sorted in ascending order.
\end{lstlisting}
\end{simbox}
\begin{answerbox}{Turn 2 [ANSWER]}
\begin{lstlisting}
def candidate(lst):
    return lst == sorted(lst)   # misses duplicate rule entirely
\end{lstlisting}
\end{answerbox}

\medskip
\noindent\textbf{Trained policy $\dbudget{=}1.0$} (pass@1~$=1.00$, 2 questions):

\begin{askbox}{Turn 1 [ASK]}
\begin{lstlisting}
1. What is the specific order the list should be sorted in?
2. How should the function handle lists with duplicate numbers?
\end{lstlisting}
\end{askbox}
\begin{simbox}
\begin{lstlisting}
1. The list should be sorted in ascending order.
2. The function should return False if the list has more than
   1 duplicate of the same number.
\end{lstlisting}
\end{simbox}
\begin{answerbox}{Turn 2 [ANSWER]}
\begin{lstlisting}
def candidate(lst):
    if lst != sorted(lst):
        return False
    from collections import Counter
    counts = Counter(lst)
    return all(c <= 2 for c in counts.values())
\end{lstlisting}
\end{answerbox}

The baseline resolves the ambiguous sort direction but never reaches the
truncated duplicate rule, producing a partially correct solution.
The trained policy recognises both signals of missing information: the
unspecified sort direction \emph{and} the visibly cut-off duplicate sentence
(\texttt{``If list has certain duplicate of the same number,''}), and
addresses both in a single two-part question, recovering the complete
specification.
This directly illustrates why Figure~\ref{fig:deg_group_qs} shows higher mean
question counts on harder multi-degradation problems.

\section{System Prompts}
\label{sec:appendix_prompts}

The policy receives a minimal zero-shot system prompt instructing it to respond
with either \texttt{[ASK]} followed by a clarifying question or
\texttt{[ANSWER]} followed by Python code, with one action permitted per turn.
We deliberately avoided few-shot examples in the policy prompt: providing
demonstrations would risk biasing the policy toward particular question styles
or strategies, confounding the learned behaviour with the choice of examples.
The GPT-4o-mini oracle simulator receives a separate system prompt that grants
it access to the full ground-truth specification and instructs it to answer
only what is asked, without volunteering test cases or the solution.

The placeholder \texttt{\{original\_prompt\}} in the oracle simulator prompt
is filled at runtime with the ground-truth un-degraded problem specification.

\clearpage

\begin{figure}[t]
\begin{promptbox}{Policy system prompt}
\begin{lstlisting}
You are a coding assistant. Given a coding task below, you must either:
  - Ask a clarifying question by responding with [ASK] followed by
    your question.
  - Write your Python solution by responding with [ANSWER] followed
    by the code.
Important rules:
  - Respond with ONLY one action per turn ([ASK] or [ANSWER]).
  - When you have enough information, write the code.
  - Do not explain your reasoning -- just output the action directly.
\end{lstlisting}
\end{promptbox}
\end{figure}

\begin{figure}[H]
\begin{promptbox}{Oracle simulator system prompt (count mode)}
\begin{lstlisting}
You are a helpful assistant who holds the complete specification
for a coding problem.

Full specification:
{original_prompt}

Note: The agent may refer to the function by a different name
(e.g., "candidate"). Treat any function name the agent uses as
referring to the function described above.

Rules:
1. Answer ONLY the specific question(s) the agent asks.
   Do not volunteer extra information.
2. Do not reveal test cases or the full solution.
3. Keep your answer brief and factual.
4. After your answer, on a new line write EXACTLY:
   QUESTION_COUNT: N
   where N is the number of distinct atomic questions you identified
   in the agent's message.

Counting rules for N:
- Each distinct piece of information being requested = 1.
- "What is X and what is Y?" = 2.
- "What is X?" = 1.
- Conjunctions like "and", "also", "additionally" between separate
  requests = separate questions.
- "Describe the full behavior of X" where X is one concept = 1.
\end{lstlisting}
\end{promptbox}
\end{figure}

\end{document}